\documentclass[sigconf]{acmart}
\usepackage{adjustbox}
\usepackage{balance}

\begin{document}

\copyrightyear{2019} 
\acmYear{2019} 
\setcopyright{acmcopyright}
\acmConference[EMDL'19]{The 3rd International Workshop on Deep Learning for Mobile Systems and Applications}{June 21, 2019}{Seoul, Republic of Korea}
\acmBooktitle{The 3rd International Workshop on Deep Learning for Mobile Systems and Applications (EMDL'19), June 21, 2019, Seoul, Republic of Korea}
\acmPrice{15.00}
\acmDOI{10.1145/3325413.3329790}
\acmISBN{978-1-4503-6771-4/19/06}

\title[ActiveHARNet: Towards On-Device Deep Bayesian Active Learning for HAR]{ActiveHARNet: Towards On-Device Deep Bayesian Active Learning for Human Activity Recognition}

\author{Gautham Krishna Gudur}
\affiliation{%
  \institution{Global AI Accelerator, Ericsson}}
\email{gautham.krishna.gudur@ericsson.com}

\author{Prahalathan Sundaramoorthy}
\affiliation{
  \institution{University of Southern California}}
\email{prahalat@usc.edu}

\author{Venkatesh Umaashankar}
\affiliation{%
  \institution{Ericsson Research}}
\email{venkatesh.u@ericsson.com}

\renewcommand{\shortauthors}{Gautham Krishna Gudur et al.}

\sloppy

\begin{abstract}
Various health-care applications such as assisted living, fall detection etc., require modeling of user behavior through Human Activity Recognition (HAR). HAR using mobile- and wearable-based deep learning algorithms have been on the rise owing to the advancements in pervasive computing. However, there are two other challenges that need to be addressed: first, the deep learning model should support on-device incremental training (model updation) from real-time incoming data points to learn user behavior over time, while also being resource-friendly; second, a suitable ground truthing technique (like Active Learning) should help establish labels on-the-fly while also selecting only the most informative data points to query from an oracle. Hence, in this paper, we propose \textit{ActiveHARNet}, a resource-efficient deep ensembled model which supports on-device Incremental Learning and inference, with capabilities to represent model uncertainties through approximations in Bayesian Neural Networks using dropout. This is combined with suitable acquisition functions for active learning. Empirical results on two publicly available wrist-worn HAR and fall detection datasets indicate that \textit{ActiveHARNet} achieves considerable efficiency boost during inference across different users, with a substantially low number of acquired pool points (at least 60\% reduction) during incremental learning on both datasets experimented with various acquisition functions, thus demonstrating deployment and Incremental Learning feasibility.
\end{abstract}



\maketitle

\section{Introduction}
Human Activity Recognition (HAR) is an important technique to model user behavior for performing various health-care applications such as fall detection, fitness tracking, health monitoring, etc. The significant escalation in usage of mobile and wearable devices has opened up multiple venues for sensor-based HAR research with various machine learning algorithms. Until recently, machine learning and deep learning algorithms for HAR have been restricted to cloud/server and Graphics Processing Units (GPUs) for obtaining good performance. However, this paradigm has begun to shift with increasing compute capabilities vested in latest smartwatches and mobile phones. Especially, on-device machine learning for monitoring physical activities has been on the rise as an alternative to server-based computes owing to communication and latency overheads \cite{survey}.

A special interest in bringing deep learning to mobile and wearable devices (incorporating deep learning on the edge) has been an active area of research owing to its automatic feature extraction capabilities, in contrast to conventional machine learning models which mandate domain knowledge to craft shallow heuristic features. One of the unexplored areas involving deep learning for such HAR tasks is \textit{Active Learning} - a technique which gives a model, the ability to learn from real-world unlabeled data by querying an oracle \cite{ActiveActivity}. The integration of Bayesian techniques with deep learning provide us a convenient way to represent model uncertainties by linking \textit{Bayesian Neural Networks (BNNs)} with Gaussian processes using \textit{Dropout} \cite{dropout_yaringal} (Section \ref{subsection:background}). These are effectively combined with contemporary deep active learning \textit{acquisition functions} for querying the most uncertain data points from the oracle (Section \ref{subsection:acquisition_functions}).

However, these works have not been considerably discussed in resource-constrained (on-device) HAR scenarios; particularly, during \textit{Incremental Learning} - where a small portion of unseen user data is utilized to update the model, thereby adapting to the new user's characteristics \cite{HARNet}. In this paper, we investigate uncertainty-based deep active learning strategies in HAR and fall detection scenarios for wearable-devices. The main scientific contributions of this paper include:
\begin{itemize}
    \item A study of a sensor-based light-weight Bayesian deep learning model across various users on wrist-worn heterogeneous HAR and Fall Detection datasets.
    \item Leveraging the benefits of \textit{Bayesian Active Learning} to model uncertainties, and exploiting several acquisition functions to instantaneously acquire ground truths on-the-fly, thereby substantially reducing the labeling load on oracle.
    \item Enabling Incremental Learning to facilitate continuous model updation on-device from incoming real-world data independent of users (\textit{User Adaptability}), in turn eliminating the need to retrain the model from scratch.
\end{itemize}

The rest of the paper is organized as follows. Section \ref{section:related} presents the related work in the area of deep learning and active learning for HAR. Section \ref{section:our_approach} discusses about our approach to model uncertainties, the Bayesian \textit{HARNet} model architecture, and the acquisition functions used for querying the oracle. The baseline evaluations for the model in an user-independent scenario on two different datasets are elucidated in Section \ref{section:baseline}. This is followed by systematic evaluation of the same with the proposed \textit{ActiveHARNet} architecture in resource-constrained incremental active learning scenarios in Section \ref{section:ActiveHARNet}.

\section{Related Work}
\label{section:related}

Technological advancements in pervasive and ubiquitous health-care have drastically improved the quality of human life, and hence has been an actively explored research area \cite{Healthcare1}, \cite{Healthcare3}. Sensor-based deep learning has been an evolving domain for computational behavior analysis and health-care research. Mobile/wearable deep learning for HAR and fall detection \cite{wearable_fall} among elders have become the need of the hour for patient monitoring; they have been widely used with promising empirical results by effectively capturing the most discriminative features using convolutional \cite{iot_wearable} and recurrent neural networks, Restricted Boltzmann Machine \cite{deepwatch_wristsense} and other ensembled models \cite{HARNet}, \cite{DeepSense}. However, these works assume that the incoming streams of data points are labeled in real-time, thereby necessitating ground truthing techniques like active learning to handle unlabeled data.

Conventional Active Learning (AL) literature \cite{Settles} mostly handle low-dimensional data for uncertainty estimations, but do not generalize to deep neural networks as the data is inherently high-dimensional \cite{BayesianAL}. Deep active learning using uncertainty representations - which presently are the state-of-the-art AL techniques for high-dimensional data, have had very sparse literature. With the advent of Bayesian approaches to deep learning, Gal et al. \cite{BayesianAL} proposed Bayesian Active Learning for image classification tasks, and is proven to learn from small amounts of unseen data, while Shen et al. \cite{NER_AL} incorporate similar techniques for NLP sequence tagging. However, these techniques are predominantly not discussed for sensor and time-series data.

Incorporating AL for obtaining ground truth in mobile sensing systems has been addressed in few previous works discussed as follows. Hossain et al. \cite{ActiveActivity} incorporate a dynamic k-means clustering approach with AL in HAR tasks. However, this work was before the ubiquitousness of deep learning algorithms, thereby making the model dependent on heuristic hand-picked features. Lasecki et al. \cite{Legion} discuss about real-time crowd sourcing on-demand to recognize activities using Hidden Markov Models (HMMs) on videos, but not on inertial data. Moreover, deep learning models have vastly outperformed HMMs in video classification setting. Bhattacharya et al. \cite{SouravUnlabeled} propose a compact and sparse coding framework to reduce the amount of ground truth annotation using unsupervised learning. Although these works seem to achieve impressive results, the feasibility of on-device Incremental Learning (model updation) scenarios with unlabeled data still seems debatable.

In this paper, we propose \textbf{\textit{ActiveHARNet}}: a unified and novel deep Incremental Active Learning framework for Human Activity Recognition and Fall Detection tasks for ground truthing on resource-efficient platforms, by modeling uncertainty estimates on deep neural networks using Bayesian approximations, thereby efficiently adapting to new user behavior.

\section{Our Approach}
\label{section:our_approach}

\begin{figure}[ht]
\centering
	\includegraphics[width=\linewidth]{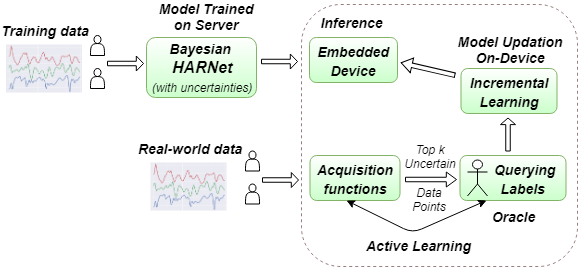}
\caption{\textit{ActiveHARNet} Architecture}
\label{fig:Block_Diagram}
\end{figure}

In this section, we discuss in detail about our proposed \textit{ActiveHARNet} pipeline/architecture (showcased in Figure \ref{fig:Block_Diagram}), and our approach to perform Incremental Active Learning.

\subsection{Background on Modeling Uncertainties}
\label{subsection:background}

Bayesian Neural Networks (BNNs) offer a probabilistic interpretation to deep learning models by incorporating Gaussian prior (probability distributions) - $p(\omega)$ over our model parameters (weights - $\omega$), thereby modeling output uncertainties. The likelihood model for a classification setting with c classes and x input points is given by,
$$
p(y=c | x, \omega) = softmax(f^\omega(x))
$$
where $f^\omega(x)$ is the model output.  However, the posterior distribution of BNNs are not easily tractable, hence it becomes computationally intensive for training and inference.

Gal et al. propose that, \textit{Dropout} - a stochastic regularization technique \cite{dropout}, can also perform approximate inference over a deep Gaussian process \cite{dropout_yaringal}, thereby learn the model posterior uncertainties without high computational complexities. This is equivalent to performing Variational Inference (VI), where the posterior distribution is approximated by finding another distribution $q_\theta^*(\omega)$, parameterized by $\theta$, within a family of simplified tractable distributions, while minimizing the Kullback-Leibler (KL) divergence between $q_\theta^*(\omega)$ and the true model posterior $p(\omega | D_{train})$. 

During inference, we can estimate the mean and variance of the BNN's output by applying dropout before every fully-connected layer during train and test time for multiple stochastic passes (\textit{T}). This is equivalent to obtaining predictions and uncertainty estimates respectively from the approximate posterior output of the neural network, thereby making the Bayesian NN \textit{non-deterministic} \cite{dropout_yaringal}. The predictive distribution for a new data point input $x^*$ can be obtained by,
$$
p(y^* | x^*, D_{train}) = \int p(y^* | x^*, \omega) p(\omega | D_{train}) d \omega
$$
where $p(\omega | D_{train}) = q_\theta^*(\omega)$, and $q_\theta^*(\omega)$ is the dropout distribution approximated using VI. Dropout, being a light-weight operation in most existing NN architectures, enables easier and faster approximation of posterior uncertainties.

\subsection{Model Architecture}
\label{subsection:architecture}

\begin{figure}[ht]
  \centering
    \begin{tabular}{@{}c@{}}
      \includegraphics[width=\linewidth, height=130pt]{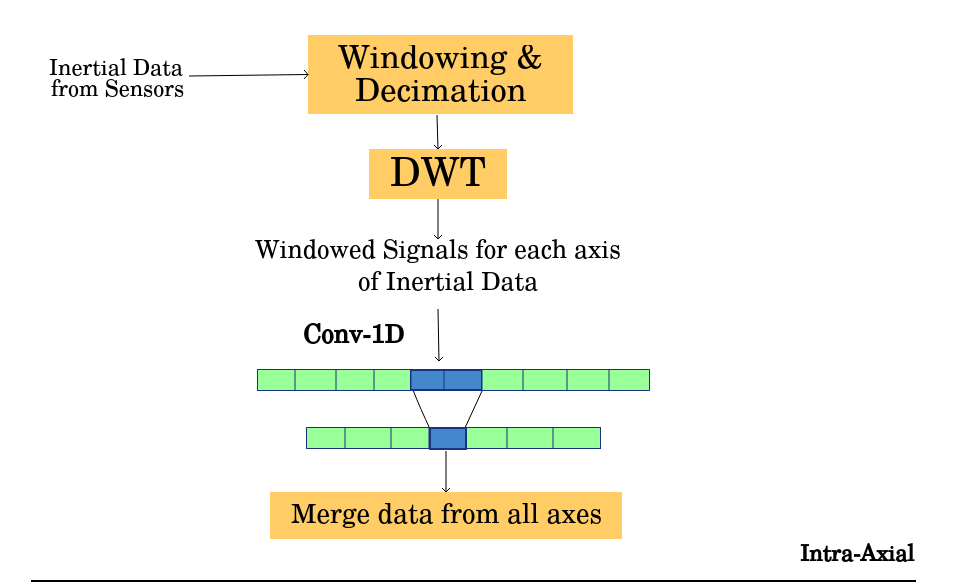} 
    \end{tabular}
    \begin{tabular}{@{}c@{}}
      \includegraphics[width=\linewidth, height=120pt]{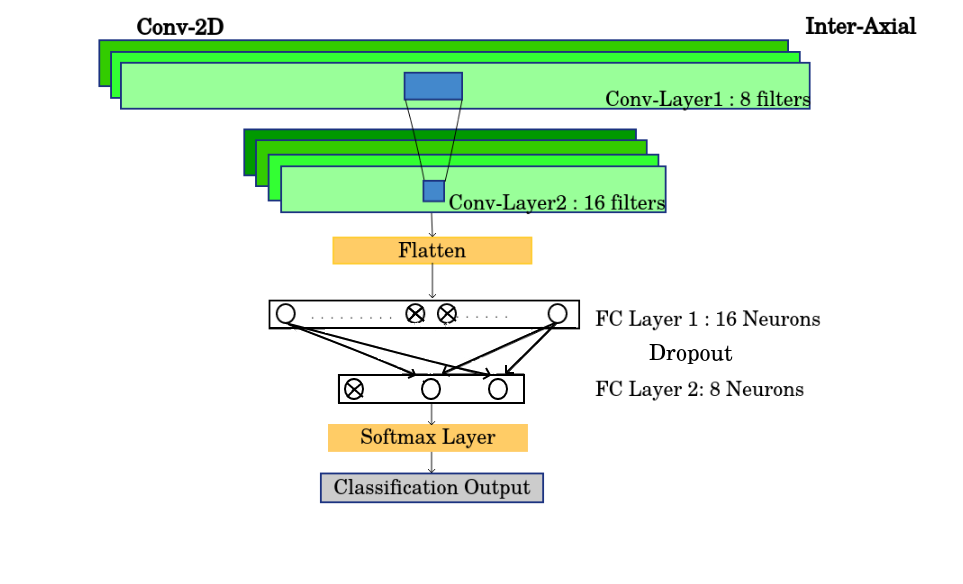} 
    \end{tabular}
  \caption{\textit{HARNet Architecture}}
  \label{fig:HARNet_Architecture}
\end{figure}

To extract discriminative features from inertial data, a combination of local features and spatial interactions between the three axes of the accelerometer can be exploited, using a combination of convolutional 1D and 2D layers.

\textit{Intra-Axial dependencies} are captured using a two-layer stacked convolutional 1D network, with 8 and 16 filters each and kernel size 2. Batch normalization is performed followed by a max-pooling layer of size 2.

\textit{Inter-axial dependencies} from the concatenated intra-axial features are captured using a two-layer stacked 2-D CNN network comprising of 8 and 16 filters each with receptive field size 3x3, followed by batch normalization and a max pooling layer of size 3x2.

This is followed by two Fully-Connected (FC) layers with 16 and 8 neurons each, with weight regularization (L2-regularizer with a weight decay constant), and ReLU activations. A dropout layer with a probability of 0.3 is applied, followed by a softmax layer to get the probability estimates (scores). The categorical-cross entropy loss of the model is minimized using Adam optimizer with a learning rate of $2e^{-4}$ and implemented using the TensorFlow framework. We choose this \textit{HARNet} architecture \cite{HARNet} with extensive parametric optimization, as it is a state-of-the-art architecture for heterogeneous HAR tasks by taking into account the efficiency, model size and inference times, with extremely less parameters ($\sim$31,000 parameters) compared to contemporary deep learning architectures.

In order make \textit{HARNet} (Figure \ref{fig:HARNet_Architecture}) a Bayesian NN so as to obtain uncertainty estimates, we introduce a standard Gaussian prior on the set of our model parameters. Also, to perform approximation inference in our model, we perform dropout at train and test-time as discussed in Section \ref{subsection:background} to sample from the approximate posterior using multiple stochastic forward passes (\textit{MC-dropout}) \cite{dropout_yaringal}. After experimenting with multiple dropout iterations (forward passes - \textit{T}), an optimal \textit{T}=10 in utilized this paper to determine uncertainties. Effectively, \textit{HARNet is a Bayesian ensembled Convolutional Neural Network (B-CNN)} which can model uncertainties, which can be used with existing acquisition functions for AL.

\subsection{Acquisition functions for Active Learning}
\label{subsection:acquisition_functions}

As stated in \cite{BayesianAL}, given a classification model $M$, pool data $D_{pool}$ obtained from real-world, and inputs $x \in D_{pool}$, an acquisition function $a(x, M)$ is a function of $x$ that the active learning system uses to infer the next query point:
$$
    x^* = argmax_{x \in D_{pool}} a(x, M).
$$

Acquisition functions are used in active learning scenarios for approximations in Bayesian CNNs, thereby arriving at the most efficient set of data points to query from $D_{pool}$. We examine the following acquisition functions to determine the most suitable function for on-device computation:

\subsubsection{Max Entropy}
Pool points are chosen that maximize the predictive entropy \cite{Max_entropy}.
\begin{align*}
    \mathbb{H}& [y | x, D_{train}] := - \sum_c p(y=c | x, D_{train}) \log p(y=c | x, D_{train})
\end{align*}

\subsubsection{Bayesian Active Learning by Disagreement (BALD)}
In BALD, pool points are chosen that maximize the mutual information between predictions and model posterior \cite{BALD}. The points that maximize the acquisition function are the points that the model finds uncertain on average, and information about model parameters are maximized under the posterior that disagree the most about the outcome.
$$
\mathbb I[y, \omega | x, D_{train}] =
\mathbb H[y | x, D_{train}] - E_{p(\omega | D_{train})} \big[\mathbb H[y | x, \omega] \big]
$$
where $\mathbb H[y | x, \omega] $ is the entropy of $y$, given model weights $\omega$.

\subsubsection{Variation Ratios ($VR$)}
The LC (Least Confident) method for uncertainty based pool sampling is performed in $VR$ \cite{Var_Ratios}.
$$
variation-ratio[x] := 1 - \max_y p(y | x, D_{train})
$$

\subsubsection{Random Sampling}
This acquisition function is equivalent to selecting a point from a pool of data points uniformly at random.

\section{Baseline Evaluation}
\label{section:baseline}

To evaluate our Bayesian \textit{ActiveHARNet} framework on an embedded platform, we experiment and analyze the results on two wrist-worn public datasets, which were performed across multiple users in real-world. Datasets with multiple users were rigorously selected to exhibit the capabilities of \textit{ActiveHARNet} like incremental active learning and user adaptability. This is essentially the pre-training phase where the model is stocked in the embedded system. To approximate our posterior with predictive uncertainties, we test our BNN model over \textit{T}=10 stochastic iterations, and average our predictions to calculate our final efficiencies on both datasets. Also, each model is trained for a maximum of 50 epochs to establish the baseline efficiencies, as we observe that loss saturates extensively and does not converge after the same.

\subsection{Heterogeneous Human Activity Recognition (HHAR) Smartwatch Dataset}

\textit{HHAR} dataset, proposed by Allan et al. \cite{HHAR}, contains accelerometer data from different wearables - two LG G smartwatches and two Samsung Galaxy Gears across nine users performing six activities: Biking, Sitting, Standing, Walking, Stairs-Up, Stairs-Down in real-time heterogeneous conditions.

\subsubsection*{Data Preprocessing} As performed in \cite{HARNet}, we first segment the raw inertial accelerometer data into two-second non-overlapping windows. To handle disparity in sampling frequencies across devices, we perform \textit{Decimation} - a down-sampling technique on all windows, to the least sampling frequency (100 Hz - Samsung Galaxy Gear) to obtain uniform distribution in data. Hence, the size of each window ($w_a$) is 200. Further, to obtain temporal and frequency information, we perform \textit{Discrete Wavelet Transform (DWT)} and take only the Approximate coefficients. Note that, performing such operations compress the size of the sensor data by more than $\sim$50\%. Also, we utilize only accelerometer data and not gyroscope, since the former reduces the size of the dataset by half without compromising much on accuracy.

\begin{figure}[ht]
\centering
	\includegraphics[width=\linewidth, height=150pt]{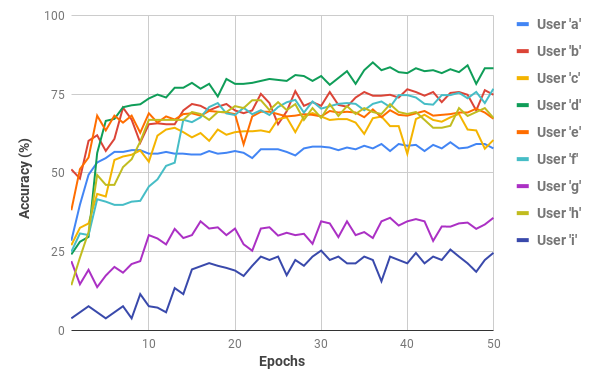}
\caption{Baseline Accuracy (\%) vs Epochs on \textit{HHAR} for Users `a' through `i'}
\label{fig:All_Accuracies}
\end{figure}

Initially, we benchmark our baseline accuracies on the server using Bayesian \textit{HARNet} with the \textit{Leave-One-User-Out (LOOCV)} strategy. The test user samples are split in random into $D_{test}$ and $D_{pool}$ points, with $D_{pool}$ slightly higher than $D_{test}$ (70-30 ratio) as an approximation of real-world incoming data, while the unseen $D_{test}$ is always used for evaluation purposes only in our experiment for both datasets. The exact number of $D_{pool}$ and $D_{test}$ differs with various users, and is subjective in real-time. The average accuracy using \textit{LOOCV} is observed to be $\sim$61\%. Also, from Figure \ref{fig:All_Accuracies}, we can infer that the model performs the best on user `d' data with $\sim$84\% classification accuracy, while the classification accuracies of user `i' and user `g' are the least with $\sim$25\% and $\sim$36\% respectively. These disparate changes in accuracies can be attributed to the unique execution style of activities by the users.

\subsection{Notch Wrist-worn Fall Detection Dataset}

This dataset uses an off-the-shelf \textit{Notch} sensor as performed in \cite{smartfall} (only the wrist-worn accelerometer data is used). The dataset is collected by seven volunteers across various age groups performing simulated falls and activities (activities are termed as not-falls).

\subsubsection*{Data Preprocessing} We segment the raw inertial data of each activity into non-overlapping windows with the given standardized sampling frequency of 31.25 Hz as in \cite{smartfall}, hence there is no decimation performed here. Further, similar to the \textit{HHAR} dataset preprocessing, we perform \textit{Discrete Wavelet Transform (DWT)} to obtain temporal and frequency information of the sensor data and take only the Approximate coefficients. Note that performing these operations compress the size of the sensor data by more than $\sim$50\%.

From Table \ref{table:notch_base}, we can observe the f1-scores and accuracies with \textit{LOOCV} strategy on the Bayesian \textit{HARNet}. f1-score would be a better estimate for handling the imbalance in falls and activities, since fall is the rare case event in this binary classification setting. A similar randomized $D_{test}$ and $D_{pool}$ split (70-30 ratio) is performed, and the average f1-score on $D_{test}$ using Bayesian \textit{HARNet} is found to be 0.927, which is substantially higher than \cite{smartfall}, whose best-performing deep learning model's f1-score is calculated to be 0.837 from its precision and recall scores.

\begin{table}[ht]
\caption{Baseline f1-scores and Accuracies on \textit{Notch}}
\label{table:notch_base}
\centering
\begin{adjustbox}{width=0.475\textwidth}
\begin{tabular}{|c|c|c|c|c|c|c|c|}
\hline
                  & \textbf{User 1} & \textbf{User 2} & \textbf{User 3} & \textbf{User 4} & \textbf{User 5} & \textbf{User 6} & \textbf{User 7} \\ \hline
\textbf{f1-score} & 0.9326          & 0.9214          & 0.9357          & 0.9372          & 0.9195          & 0.9229          & 0.9248          \\ \hline
\textbf{Accuracy} & 97.02           & 94.44           & 94.05           & 95.36           & 94.08           & 94.59           & 94.65           \\ \hline
\end{tabular}
\end{adjustbox}
\end{table}

\section{Incremental Active Learning}
\label{section:ActiveHARNet}

In order to handle ground truth labeling and incorporate model weight updation on incoming test user data, we experiment incremental active training with \textit{LOOCV} on both datasets by deploying the system on a Raspberry Pi 2. We choose this single-board computing platform, since it has similar hardware and software specifications with predominant contemporary wearable devices. The number of acquisition windows used for incremental active training from $D_{pool}$ can be governed by the \textit{acquisition adaptation factor} $\eta \in$ [0, 1]. The incremental model updation was simulated for a maximum of 10 epochs, owing to its non-convergence in loss thereafter.

\subsection{ActiveHARNet on HHAR dataset}
\label{subsection:ActiveHARNet_HHAR}

\begin{figure}[ht]
\centering
	\includegraphics[width=\linewidth, height=150pt]{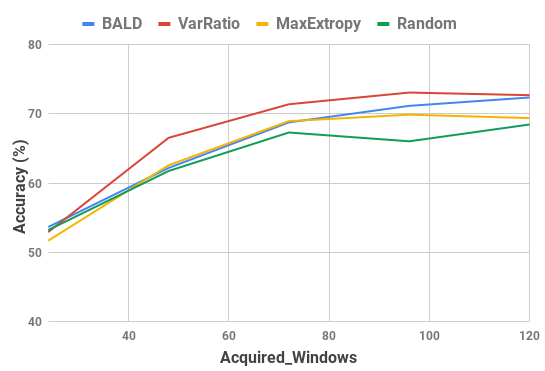}
\caption{\textit{ActiveHARNet} on \textit{HHAR} for User `i' across acquisition functions - Accuracy (\%) vs Acquisition Windows}
\label{fig:user_i}
\end{figure}

We analyze various AL acquisition functions mentioned in section \ref{subsection:acquisition_functions} for all users, particularly for the worst performing user `i' (least classification accuracy), and we can infer from Figure \ref{fig:user_i} that Variation Ratios ($VR$) acquisition function performs the best, while Random Sampling has the least classification accuracy as expected. In $VR$, only 62 ($\eta$=0.5 or 50\%) acquisition windows from the total $D_{pool}$ (123 windows) were required for user `i' to achieve a test accuracy of $\sim$70\% from a baseline accuracy of $\sim$25\%, which is a substantial increase of $\sim$45\% in test accuracy. With $\eta$=1.0 (all $D_{pool}$ windows), a maximum of $\sim$73\% is achieved.

We test \textit{ActiveHARNet} using $VR$ across all users, and observe that the average baseline accuracy ($\eta$=0) increases from 61\% to a maximum of $\sim$86\% ($\eta$=1) (Table \ref{table:hhar_active}). Also, very few acquisition windows ($\eta$=0.4, accuracy=83.05\%) from incoming $D_{pool}$ are found to be sufficient for achieving competitive efficiencies (85.87\%) as $\eta$=1.0. Note that, $\eta$=0.0 gives the efficiency of the pre-trained model without any data points acquired during incremental learning.

\begin{table}[ht]
\caption{\textit{ActiveHARNet} on \textit{HHAR} with Variation Ratios for all users - Acquisition Windows ($\eta$) vs Accuracy (\%)}
\label{table:hhar_active}
\centering
\begin{adjustbox}{width=0.475\textwidth}
\begin{tabular}{|c|c|c|c|c|c|c|c|c|c|c|}
\hline
\textbf{$\eta$} & \textbf{User a} & \textbf{User b} & \textbf{User c} & \textbf{User d} & \textbf{User e} & \textbf{User f} & \textbf{User g} & \textbf{User h} & \textbf{User i} & \textbf{Avg.} \\ \hline
\textbf{0.0}                    & 57.83           & 74.86           & 60.5            & 83.79           & 67.25           & 76.77           & 35.78           & 67.5            & 24.66           & 61              \\ \hline
\textbf{0.2}                    & 83.52           & 89.76           & 75.7            & 91.95           & 81.53           & 79.79           & 73.39           & 78.75           & 52.92           & 78.59            \\ \hline
\textbf{0.4}                    & 89.15           & 91.72           & 80.85           & 92.3            & 85.05           & 84.57           & 76.23           & 81              & 66.53           & 83.05            \\ \hline
\textbf{0.6}                    & 91.55           & 92.18           & 82.26           & 93.26           & 87.92           & 86.96           & 77.15           & 83.5            & 71.38           & 85.13            \\ \hline
\textbf{0.8}                    & 92.64           & 93.24           & 82.28           & 93.56           & 87.52           & 88.07           & 78.58           & 82.6            & 73.07           & 85.73            \\ \hline
\textbf{1.0}                    & 92.72           & 93.16           & 85.06           & 93.64           & 89.95           & 87.96           & 76.23           & 81.375          & 72.69           & 85.87            \\ \hline
\end{tabular}
\end{adjustbox}
\end{table}

\subsection{ActiveHARNet on Notch dataset}
\label{subsection:ActiveHARNet_Notch}

\begin{figure}[ht]
\centering
	\includegraphics[width=\linewidth, height=150pt]{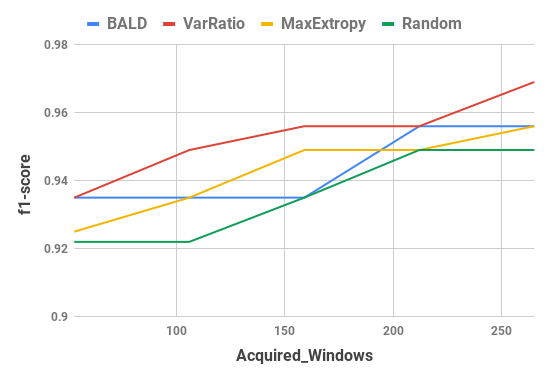}
\caption{\textit{ActiveHARNet} on \textit{Notch} for User 5 across acquisition functions - f1-scores vs Acquisition Windows}
\label{fig:notch:user_i}
\end{figure}

We analyze AL acquisition functions for all users as performed in Section \ref{subsection:ActiveHARNet_HHAR}, and showcase the f1-scores of user 5 (least performing) in Figure \ref{fig:notch:user_i}. Variation Ratios ($VR$) has the highest classification efficiency again, requiring only 150 acquisition windows ($\eta$=0.4) for a competitive f1-score of 0.956 from the total $D_{pool}$ of 265 windows ($\eta$=1.0), whose f1-score is 0.969. Random acquisition performs with the least f1-score again.

\begin{table}[ht]
\caption{\textit{ActiveHARNet} on \textit{Notch} with Variation Ratios for all users - Acquisition Windows ($\eta$) vs f1-scores}
\label{table:notch_active}
\centering
\begin{adjustbox}{width=0.47\textwidth}
\begin{tabular}{|c|c|c|c|c|c|c|c|c|}
\hline
\textbf{$\eta$} & \textbf{User 1} & \textbf{User 2} & \textbf{User 3} & \textbf{User 4} & \textbf{User 5} & \textbf{User 6} & \textbf{User 7} & \textbf{Avg.} \\ \hline
\textbf{0.0}                    & 0.932           & 0.921           & 0.936           & 0.937           & 0.92            & 0.923           & 0.925           & 0.928         \\ \hline
\textbf{0.2}                    & 0.938           & 0.924           & 0.945           & 0.947           & 0.935           & 0.932           & 0.925           & 0.935         \\ \hline
\textbf{0.4}                    & 0.943           & 0.929           & 0.961           & 0.952           & 0.949           & 0.932           & 0.932           & 0.943         \\ \hline
\textbf{0.6}                    & 0.949           & 0.929           & 0.965           & 0.952           & 0.956           & 0.945           & 0.936           & 0.948         \\ \hline
\textbf{0.8}                    & 0.943           & 0.937           & 0.968           & 0.965           & 0.956           & 0.953           & 0.942           & 0.952         \\ \hline
\textbf{1.0}                    & 0.952           & 0.937           & 0.965           & 0.956           & 0.969           & 0.945           & 0.936           & 0.951         \\ \hline
\end{tabular}
\end{adjustbox}
\end{table}

Also, from Table \ref{table:notch_active}, $VR$ across all users yield average f1-scores of 0.943 and 0.948 for $\eta$=0.4 and $\eta$=0.6 respectively, from a baseline 0.928 ($\eta$=0.0), thus scaling well to new users with substantially less data points.

\subsection{On-Device Incremental Learning}

Raspberry Pi 2 is used for evaluating incremental active learning, and an average time of $\sim$1.4 sec is utilized for each stochastic forward pass (\textit{T}) with dropout for acquisition of top $\eta$ windows. Since we perform \textit{T}=10 dropout iterations in our experiment, we observe that an average of $\sim$14 seconds are needed for querying most uncertain data points. Variation Ratios was found to be converging slightly faster with better test accuracies than BALD and MaxEntropy acquisition functions on both datasets, while Random Sampling converged relatively slower. Also, there is substantial increase in relative HHAR efficiencies during incremental active learning than that of Notch. This can be attributed to the nature of the dataset, and the different ways people perform activities. Notch, being a fall detection problem with two classes, has a higher probability of better recognition efficiencies, than a multi-class HHAR problem.

\begin{table}[ht]
\centering
\caption{Computation Time Taken for Execution per window}
\label{table:timeOnPi}
\begin{tabular}{|c|c|c|}
\hline
\multicolumn{1}{|c|}{\textbf{Process}} & \textbf{HHAR} & \textbf{Notch} \\ \hline
Inference time                  & 14 ms    & 11 ms       \\ \hline
Discrete Wavelet Transform      & 0.5 ms   & 0.39 ms         \\ \hline
Decimation                      & 3.4 ms   & $-$         \\ \hline
Time taken per epoch            & 1.8 sec  & 1.2 sec         \\ \hline
\end{tabular}
\end{table}

The model size for HHAR was $\sim315$ kB, while for Notch, it was found to be $\sim$180 kB, which is substantially small compared to conventional deep learning models. For real-time deployment feasibility, it is practical to have a threshold (upper limit) on the number of $D_{pool}$ collected at a single point of time. This can be quantified by either number of windows ($w_a$) or time taken (in seconds). We propose time as a benchmark (for instance, 15 minutes since start of an activity cycle; the setting could be personalized based on user), so that the oracle efficiently remembers the recently performed activities when queried for ground truth. Many such acquisition iterations, in turn, model updations would ideally happen during real-time, and our experiments showcased the practical feasibility of one such acquisition iteration with $\sim$14 seconds. The end-user can also have a trade-off between efficiency and model updation time, and this is proportional to the number of data points to be queried by the oracle.

\section{Conclusion}

This paper presents three new empirical contributions with emphasis on Bayesian Active Learning for embedded/wearable deep learning on HAR and fall detection scenarios. First, we benchmark our efficiencies for both datasets using the Bayesian \textit{HARNet} model, which can incorporate uncertainties. Second, we systematically analyze various acquisition functions for active learning and exploit Bayesian Neural Networks with stochastic dropout for extracting the most informative data points to be queried by the oracle, using uncertainty approximations. Third, we propose \textit{ActiveHARNet} - a resource-friendly unified framework which facilitates on-device Incremental Learning (model updation) for seamless physical activity monitoring and fall detection over-time, and can further be extended to other behavior monitoring tasks in pervasive healthcare.

\bibliographystyle{acm}
\balance
\bibliography{References}

\end{document}